\documentclass{article}
\usepackage[utf8]{inputenc}
\usepackage[a4paper, total={7.5in, 10in}]{geometry}
\usepackage{microtype}
\usepackage{graphicx}
\usepackage{subfigure}
\usepackage{amsmath}
\usepackage{mathtools}
\newcounter{todocounter}
\usepackage{todonotes}
\usepackage{amsfonts}
\usepackage{booktabs} 
\usepackage{hyperref}
\usepackage[sorting=none]{biblatex}
\usepackage[shortlabels]{enumitem}
\usepackage[title]{appendix}
\usepackage{authblk}

 \definecolor{blizzardblue}{rgb}{0.67, 0.9, 0.93}

\title{Evaluating Error Bound for Physics-Informed Neural Networks on Linear Dynamical Systems}

\author[1]{Shuheng Liu}
\author[2]{Xiyue Huang}
\author[1]{Pavlos Protopapas}
\affil[1]{Institute for Applied Computational Science, Harvard University}
\affil[2]{Carnegie Mellon University}
\date{July 2022}

\begin{document}

\maketitle

\begin{abstract}
    
    There have been extensive studies on solving differential equations using physics-informed neural networks. While this method has proven advantageous in many cases, a major criticism lies in its lack of analytical error bounds. Therefore, it is less credible than its traditional counterparts, such as the finite difference method. This paper shows that one can mathematically derive explicit error bounds for physics-informed neural networks trained on a class of linear systems of differential equations. More importantly, evaluating such error bounds only requires evaluating the differential equation residual infinity norm over the domain of interest. Our work shows a link between network residuals, which is known and used as loss function, and the absolute error of solution, which is generally unknown. Our approach is semi-phenomonological and independent of knowledge of the actual solution or the complexity or architecture of the network. Using the method of manufactured solution on linear ODEs and system of linear ODEs, we empirically verify the error evaluation algorithm and demonstrate that the actual error strictly lies within our derived bound.
    
\end{abstract}


\section{Introduction}


Differential equations play an essential role in science, engineering, as well as a wide range of mathematical modeling processes. In many cases, however, analytical solutions are nonexistent, and therefore many numerical methods (finite difference, finite volute, finite element, spectral method) have been studied  for these equations. Recently, massive attention has been paid to solving differential equations with neural networks, which is capable of learning a continuous solution while occupying relatively less storage. These networks learn to minimize the squared residual of some differential equations, which are usually believed to govern some physical systems. 

Raissi et al use the term \textit{physics-informed neural networks} (PINNs) to refer to networks that are trained on both the differential equation residuals (in an unsupervised manner) and observed data (in a supervised manner) \cite{raissi2019physics}. However, in this paper, we only consider networks trained to minimize differential equation residuals subject to initial/boundary conditions, without any data from observations. Nonetheless, we still refer to them as physics-informed neural networks  as they are only deprived of observed data while remaining informed of the underlying physics.

Unlike for tradition numerical methods, little effort has been made to ascertain the error bound of neural network solutions to differential equations. Therefore, the reliability of neural network solutions are usually questionable, making them hardly acceptable for real-world applications. Additionally, there is little interpretation for the loss function used for guiding the training process, other than that it should be as close to zero as possible. However, the link between the loss value (typically an approximation of the $L_2$ norm of ODE residual) and the error bound has not been formulated. 

In this paper, we present a mathematical proof that explicitly bounds the absolute/relative error of a solution of a class of linear dynamical systems with neural networks. 
The method makes no assumption on the network architecture or whether the network is sufficiently trained. 
Apart from the characteristics and structure of the dynamical systems in question, the error bound we derived ($O(\varepsilon t^m)$) only depends on time $t$ and the largest differential equation residual ($\varepsilon$) (infinity norm) over the domain of interest. 
We further present that, for strictly stable systems, one can derive a bound ($O(\varepsilon)$) that is independent of time $t$. 
Finally, we present a technique to tighten the error bound by dividing the time domain into subintervals and evaluating the maximum residual on each one.

Until more straightforward error evaluation approach is identified, our method can serve as a practical way to estimate the absolute error of neural network, which is generally unknown, using the network residual, which is easily computed and used as the loss function.

Our main contributions are threefold. First, we propose an algorithm for rigorous error bound evaluation. Second, we extend the algorithm to more complicated linear cases, including linear systems and equation with nonconstant coefficients. Third, we demonstrated the effectiveness of the evaluation algorithm using method of manufactured solution. 

In Section \ref{section:background}, we introduce the background of physics-informed neural networks (PINNs) and their application in solving differential equations. In Section \ref{section:previous-work}, we discuss the previous work in this field. In Section \ref{section:error-bound-proof-and-methodology}, we propose and prove a method to efficiently evaluate and bound the error. In Section \ref{section:experiments}, we show experimental results that verify our proposed method. Finally, we conclusively summarize our work and provide direction for future work in Sections \ref{section:conclusion} and \ref{section:future-work}.

\section{Background} \label{section:background}
Neural networks were first introduced as universal function approximators that learn nonlinear mappings for supervised learning tasks \cite{hornik1989multilayer}. Lagaris et al. first proposed solving differential equations using neural networks \cite{lagaris1998artificial} due to the smoothness of neural networks with appropriate activation functions. 

One main advantage of using neural networks in solving differential equations is its potential to learn the structure of the solution space and give a bundle of solutions $\mathbf{u}(t; \theta_1, \dots, \theta_n)$ where $\theta_1, \dots, \theta_n$ parametrize the differential equation and/or initial condition \cite{flamant2020solving} \cite{DesaiShaan2021OTLo}. For traditional numerical methods, a new solution must be calculated from scratch whenever any parameter in equation or initial condition changes. Neural networks, however, can in theory take in $t, \theta_1, \dots, \theta_n$ as its input and learn the structure of the solution space parametrized by $\theta_1, \dots, \theta_n$. 

Multiple approaches have been proposed to enforce initial/boundary conditions on network solutions \cite{lagaris1998artificial}  \cite{lagaris2000neural} \cite{mcfall2009artificial} \cite{lagari2020systematic} \cite{sukumar2021exact}. In this paper, we use a slightly modified version of Lagaris' reparametrization technique $\mathbf{u}(t) = (t-t_0) \mathbf{NN}(t) + \mathbf{u}_0$ to ensure $\mathbf{u}(t_0) =\mathbf{u}_0$ where $\mathbf{NN}$ is the network. Unless otherwise specified, we refer to $\mathbf{u}(t)$ as the \textit{network solution} which is already reparamtrized to satisfy the initial condition. 

To train a network solution $\mathbf{u}(t)$ for a differential equation $\mathcal{L} \mathbf{u} = \mathbf{f}$, we essentially minimize an approximation of the $L_2$ norm of differential equation residual on a domain $\Omega$

$$ \int_{I} \left( \mathcal{L} \mathbf{u} - \mathbf{f}\right)^2 \mathrm{d}t \approx \frac{|I|}{N} \sum_{\substack{i = 1\\ t_i \in I}}^{N} \left(\mathcal{L} \mathbf{u}(t_i) - \mathbf{f}(t_i) \right)^2 := \mathrm{Loss}. $$
where $\mathcal{L}$ is a (possibly nonlinear) differential operator.

Effort has also been made to redefine the loss function by including energy constraints or using the weak formulation of equation \cite{mattheakis2020hamiltonian} \cite{yu2017deep} \cite{sirignano2018dgm}. By sampling the training points $\{t_i\}_{i=1}^{N} \subset I$ in each epoch, we aim to achieve a residual norm of zero after sufficient training. The sampling strategy is also an active field of research \cite{parwani2021adversarial}.  However, the universal approximation theorem relies on the hypothesis of a sufficiently large network architecture and only asserts the existence of a set of desired weights. In the context of PINNs, two more relevant questions are 1) how to interpret the residual and 2) how to bound the error between the network solution and the exact solution using the residuals.

Little effort has been to study the failure modes and absolute error of network solutions until recent years \cite{graf2021uncertainty} \cite{guo2020energy} \cite{krishnapriyan2021characterizing}. In our work, we mathematically derive the error bound for a class of linear ODEs, which can be efficiently computed using only on the infinite norm of ODE residuals on the domain of interest. We also present a way to tighten the derived error bound by dividing the domain into smaller subintervals.

\section{Previous Work}\label{section:previous-work}

In \cite{de2021error}, Ryck and Mishra established a foundation and rationale for error of PINNs in approximating PDEs. Making Kolmogorov PDEs as an example, they have shown that there exists PINNs, approximating these PDEs such that the resulting generalization error and the total error can be made arbitrarily small. However, the existence of such neural networks does not guarantee network training converges in practice. A more practical concern is how to evaluate the error given any network (possibly illy trained) on certain equations. 

\section{Error Bound Proof and Methodology} \label{section:error-bound-proof-and-methodology}

Throughout this section, we use $\mathbf{u}: I \to \mathbb{C}^n$ to denote the neural network solution to $\mathcal{L}\mathbf{u} = \mathbf{f}$ where $I$ can be any of the forms $(t_0, t_1]$, $(t_0, t_1)$ or $(t_0, \infty)$, and $\mathcal{L}$ is a linear differential operator. In the one-dimensional case, we use non-bold font $u$ and $f$ instead of $\mathbf{u}$ and $\mathbf{f}$. The solution residual is defined as $R\mathbf{u}(t) := \mathcal{L}\mathbf{u}(t) - \mathbf{f}(t)$. The exact solution $\mathbf{u}^*(t)$ satisfies $R\mathbf{u}^*(t) \equiv \mathbf{0}$ and the exact natural response $\mathbf{u}_n^*(t)$ is defined to satisfy the associated homogeneous equation $\mathcal{L} \mathbf{u}_n^*(t) = \mathbf{0}$. Both $\mathbf{u}^*$ and $\mathbf{u}_n^*$ satisfy the same initial condition $\mathbf{u}^*(t_0) = \mathbf{u}_n^*(t_0) = \mathbf{u}_0^*$. We shall assume that each component of $\mathbf{u^*_0}$ is nonzero.

\subsection{First-Order Linear ODE with Constant Coefficients} \label{section:1st-order}

It is well known that the most general form of first-order linear ODE with constant coefficients is $u'(t) + cu(t) = f(t)$ where $c \in \Bbb C$ is a constant and $u'$ is the derivative of $u$.

\textbf{Proposition} \quad 
If the residual $Ru(t)$ of equation $u' + (\lambda + i\omega)u = f$, where $\lambda, \omega \in \mathbb{R}$, is bounded by $\varepsilon \geq 0$ on $I$, namely,

\begin{equation}\label{eq:1st-order-ode}
    \left|u' + (\lambda + i\omega) u -f\right| \leq \varepsilon \quad \forall t \in I,
\end{equation} and the network solution $u$ satisfies initial condition with $u(t_0) = u_0^* \neq 0$, then,
\begin{enumerate}[a)]
    \item The absolute error is bounded by $\left|u - u^*\right| \leq \cfrac{\varepsilon}{\lambda} \leq O(\varepsilon)$ on $I$ if the natural response $u_n^*$ is convergent ($\lambda > 0$);
    
    \item The relative error w.r.t. $u_n^*$ is bounded by $\left|\cfrac{u - u^*}{u_n^*}\right| \leq \cfrac{\varepsilon}{-\lambda |u_0^*|} \leq O(\varepsilon)$ on $I$ if the natural response $u_n^*$ is divergent ($\lambda < 0$); and

    \item The absolute and relative errors are bounded by $\left|u - u^*\right| \leq O(\varepsilon t)$ and $\left|\cfrac{u - u^*}{u_n^*}\right| \leq O(\varepsilon t)$ on $I$ if $\lambda = 0$.
\end{enumerate}

\textbf{Proof} \quad 
Multiply the integrating factor $e^{\lambda t + i\omega t}$ on both sides of Eq. \ref{eq:1st-order-ode} and evaluate the integral on $(t_0, t) \subseteq I$,

\begin{multline}
    \left| \int_{t_0}^{t} e^{\lambda \tau + i\omega \tau} \Big(u'(\tau) + (\lambda + \omega i) u(\tau) -f(\tau)\Big)\mathrm{d}\tau \right| \\
    \leq \int_{t_0}^{t} \left| e^{\lambda \tau + i\omega \tau} \Big(u'(\tau) + (\lambda + \omega i) u(\tau) -f(\tau)\Big)\right|\mathrm{d}\tau 
    \leq \int_{t_0}^{t}\left| e^{\lambda \tau + i\omega \tau} \right|\varepsilon \mathrm{d}\tau
\end{multline}
The first part of inequality holds because modulus of integral is smaller than integral of modulus. The second part holds by multiplying $e^{\lambda t + i\omega t}$ on both sides of Eq. \ref{eq:1st-order-ode} and taking the integral on $(t_0,t)$, both of which preserve inequality property.
\begin{equation}
    \left|e^{\lambda t + i \omega t} u(t) - e^{\lambda t_0 + i \omega t_0} u(t_0) - \int_{t_0}^{t}e^{\lambda \tau + i\omega \tau}f(\tau) \mathrm{d}\tau \right| \leq \varepsilon \int_{t_0}^{t} e^{\lambda \tau}\mathrm{d}\tau \\
    \label{eq:1st-order-2}
\end{equation}
L.H.S. is reduced using $\displaystyle{\int_{t_0}^{t} e^{\lambda \tau + i\omega\tau}\left(u' + (\lambda+ i \omega)u\right) \mathrm{d} \tau= \int_{t_0}^{t} \mathrm{d}\left(e^{\lambda \tau + i\omega\tau} u(\tau)\right) = \left[e^{\lambda \tau + i\omega\tau} u(\tau)\right]_{t_0}^{t}}$ and R.H.S. is reduced using $\left|e^{\lambda \tau + i\omega \tau}\right| \equiv e^{\lambda \tau}$.
\begin{equation}
    \left|u(t) - e^{\lambda \left(t_0 - t\right) + i \omega \left(t_0 - t\right)} u(t_0) - e^{-\lambda t - i\omega t}\int_{t_0}^{t}e^{\lambda \tau+i\omega \tau}f(\tau) \mathrm{d}\tau \right| \leq \varepsilon e^{-\lambda t} \int_{t_0}^{t} e^{\lambda \tau}\mathrm{d}\tau
\end{equation}
Both sides are divided by $\left|e^{\lambda t + i\omega t}\right|$.

Notice that the analytical solution is given by $$ u^*(t) = e^{\lambda (t_0 - t) + i\omega (t_0 - t)}u^*_0 + e^{-\lambda t - i\omega t} \int_{t_0}^{t} e^{\lambda \tau + i\omega \tau} f(\tau) \mathrm{d}\tau.$$

Define the alternative solution to Eq. \ref{eq:1st-order-ode} under perturbed initial condition, $u(t_0)$ as $$\widetilde{u}(t) := e^{\lambda (t_0 - t) + i\omega (t_0 - t)}u(t_0) + e^{-\lambda t - i\omega t} \int_{t_0}^{t} e^{\lambda \tau  + i\omega \tau} f(\tau) \mathrm{d}\tau.$$
With this, Eq. \ref{eq:1st-order-2} can be rewritten as 
\begin{equation}
    \left|u(t) - \widetilde{u}(t) \right| \leq \varepsilon e^{-\lambda t} \int_{t_0}^{t} e^{\lambda \tau}\mathrm{d}\tau. 
\end{equation}

By the triangle inequality, 
\begin{equation} \label{eq:1st-order-triangle-inequality}
    \left| u(t) - u^*(t) \right| \leq \left| u(t) - \widetilde{u}(t) \right| + \left| \widetilde{u}(t) - u^*(t)\right| \leq \varepsilon e^{-\lambda t} \int_{t_0}^{t} e^{\lambda \tau}\mathrm{d}\tau + \left| \widetilde{u}(t) - u^*(t)\right|.
\end{equation}

As $\widetilde{u} = u^*$ when $u(t_0) = u_0^*$, Eq. \ref{eq:1st-order-triangle-inequality} is reduced to
\begin{equation} \label{eq:1st-order-generic}
    \left| u(t) - u^*(t) \right| \leq \varepsilon e^{-\lambda t} \int_{t_0}^{t} e^{\lambda \tau}\mathrm{d}\tau.
\end{equation}

If $\lambda > 0$, Eq. \ref{eq:1st-order-generic} gives rise to the absolute error bound
\begin{equation} \label{eq:1st-order-abs}
    \left| u(t) - u^*(t) \right| \leq \varepsilon \frac{1 - e^{\lambda (t_0 - t)}}{\lambda} \leq \frac{\varepsilon}{\lambda} = O(\varepsilon) \quad (\lambda > 0).
\end{equation}

If $\lambda < 0$, dividing Eq. \ref{eq:1st-order-generic} by $ \left|u_n^*(t)\right| = \left|e^{\lambda (t_0 -t) + i\omega (t_0 - t) } u_0^*\right| = e^{\lambda (t_0 -t)}|u_0^*| $ yields the relative error bound
\begin{equation}
    \left| \frac{u(t) - u^*(t)}{u_n^*(t)} \right| \leq \varepsilon \frac{e^{-|\lambda| (t - t_0)} - 1}{-|\lambda| |u_0^*|} = \varepsilon \frac{1- e^{-|\lambda| (t - t_0)}}{|\lambda| |u_0^*|}  \leq \frac{\varepsilon}{|\lambda| |u_0^*|} = O(\varepsilon) \quad (\lambda < 0).
\end{equation}

If $\lambda = 0$, the integral on R.H.S. of Eq. \ref{eq:1st-order-generic} is reduced to $(t - t_0)$, and therefore the absolute error bound is
\begin{equation}\label{eq:1st-order-oscillate}
    \left| u(t) - u^*(t) \right| \leq \varepsilon(t - t_0) = O(\varepsilon t) \quad (\lambda  = 0).
\end{equation}
Since the natural response has constant modulus $\left|u_n^*(t)\right| = \left|e^{i\omega(t_0 - t)}u_0^*\right| \equiv \left|u_0^*\right|$ when $\lambda = 0$, the relative error with respect to the natural response is bounded by $O(\varepsilon t)$ as well.
\subsection{Higher-Order Linear ODE with Constant Coefficients}\label{section:higher-order}
\textbf{Proposition} \quad 
Let the residual $Ru(t)$ of the higher-order equation $u^{(n)} + a_{n-1}u^{(n-1)} + \dots + a_0 u = f$ be bounded by some $\varepsilon \geq 0$ on $I$, where $u^{(n)}$ is the $n$-th order derivative of $u$, namely,

\begin{equation} \label{eq:higher-order-ode}
    \left|u^{(n)} + a_{n-1}u^{(n-1)} + \dots + a_0 u - f\right| \leq \varepsilon \quad \forall t \in I.
\end{equation}

Let the network solution $u$ satisfy initial conditions $u^{(k)}(t_0) = u_0^{*(k)}$ ($k=0,\dots,n-1$). By the fundamental theorem of algebra, the characteristic polynomial $p_{c}(x)$ can be uniquely factorized as
\begin{equation} \label{eq:higher-order-polynomial}
    p_{c}(x) := x^n + a_{n-1}x^{n-1} + \dots + a_0 = \prod_{k=0}^{n-1}(x + \lambda_k + i\omega_k).
\end{equation}
It is well-known that the exact solution has the form $u^*(t) = u_p^*(t) \sum\limits_{k=0}^{n-1} c_k \exp(\lambda_k t + i \omega_k t)$, where $u_p^*$ is any particular solution to the original equation and $c_0, \dots, c_{n-1}$ are constants chosen to satisfy the initial conditions.

Let $m$ be the total number of $k$ in Eq. \ref{eq:higher-order-polynomial} such that $\lambda_k = 0$, then the absolute error is bounded by: 

\begin{equation} 
\left| u - u^* \right| \leq O(\varepsilon t^m)  \,\,\, {\rm if} \,\,\, \lambda_k \geq 0  \,\,\, {\rm for \,\, all} \,\,\, k.
\end{equation}


\textbf{Proof} \quad
For brevity, we prove the second-order case here to provide an intuition of the complete proof, which is presented in Appendix \ref{appendix:higher-order-proof}.

In the second-order case, Eq. \ref{eq:higher-order-ode} can be reduced to 

\begin{equation}\label{eq:2nd-order-factorized}
    \left|u'' + \left(\lambda_1 + i\omega_1 + \lambda_2 +i\omega_2\right) u' + \left(\lambda_1 + i\omega_1\right)\left(\lambda_2 + i\omega_2\right)u - f\right| \leq \varepsilon \quad (\lambda_1 \geq \lambda_2),
\end{equation}
or, equivalently,
\begin{equation}\label{eq:2nd-order-factorized-2}
    \left|\Big(u' + \left(\lambda_1 + i\omega_1\right)u\Big)' + \left(\lambda_2 +i\omega_2\right)\Big(u' + \left(\lambda_1 + i\omega_1\right)u\Big) - f\right| \leq \varepsilon.
\end{equation}
Let $v = u' + (\lambda_1 + i\omega_1)u$, Eq. \ref{eq:2nd-order-factorized-2} is then reduced to a first-order inequality w.r.t. $v$
\begin{equation} \label{eq:2nd-order-reduced}
    \left|v' + \left(\lambda_2 + i\omega_2 \right)v -f\right| \leq \varepsilon.
\end{equation}
By Eq. \ref{eq:1st-order-generic},
\begin{equation} \label{eq:2nd-order-reduced-bound}
    \left| v(t) - v^*(t) \right| \leq \varepsilon e^{-\lambda_2 t} \int_{t_0}^{t} e^{\lambda_2 \tau}\mathrm{d}\tau,
\end{equation}
where $v^*(t) = {u^{*}}'(t) + (\lambda_1 + i\omega_1) u^*(t)$. Substituting $v = u' + (\lambda_1 + i\omega_1)u$ into Eq. \ref{eq:2nd-order-reduced-bound} yields
\begin{equation} \label{eq:2nd-order-bound-2}
    \left| u'(t) + \left(\lambda_1 + i\omega_1\right)u(t) - v^*(t) \right| \leq \varepsilon e^{-\lambda_2 t} \int_{t_0}^{t} e^{\lambda_2 \tau}\mathrm{d}\tau = \varepsilon\frac{1-e^{\lambda_2 (t_0 - t)}}{\lambda_2}
\end{equation}
Multiplying Eq. \ref{eq:2nd-order-bound-2} by $e^{\lambda_1 t + i\omega_1 t}$, taking the integral on $(t_0, t) \subseteq I$, and dividing by $\left|e^{\lambda_1 t + i\omega_1 t}\right|$, we have
\begin{equation}\label{eq:2nd-order-bound-master}
    \left|u(t) - u^*(t)\right| \leq \varepsilon \frac{1}{\lambda_1\lambda_2}\left(1 - \frac{\lambda_1e^{-\lambda_2 t}-\lambda_2e^{-\lambda_1 t}}{\lambda_1 - \lambda_2}\right) =: \varepsilon\phi(t;\lambda_1, \lambda_2)
\end{equation}
If $\lambda_1, \lambda_2 > 0$, it can be verified that $\phi(t;\lambda_1, \lambda_2)$ is strictly increasing on $I$ and is bounded by $\left[0, \cfrac{1}{\lambda_1\lambda_2}\right)$. Therefore 
\begin{equation}
    \left|u(t) - u^*(t)\right| \leq \cfrac{\varepsilon}{\lambda_1\lambda_2} = O(\varepsilon)
\end{equation}

If $\lambda_1 > \lambda_2 = 0$, taking the limit $\lambda_2 \to 0$ in Eq. \ref{eq:2nd-order-bound-master}, there is
\begin{equation} \label{eq:2nd-order-lambda2=0}
\left|u(t) - u^*(t)\right| \leq \lim_{\lambda_2 \to 0}\varepsilon\phi(t;\lambda_1, \lambda_2) =  \frac{\varepsilon}{\lambda_1^2} \left(e^{-\lambda_1 t} + \lambda_1 t - 1\right) \leq \cfrac{\varepsilon t}{\lambda_1}= O(\varepsilon t).
\end{equation}

If $\lambda_1 = \lambda_2 = 0$, taking the double limit $\lambda_1, \lambda_2 \to 0$ in Eq. \ref{eq:2nd-order-bound-master}, there is
\begin{equation}\label{eq:2nd-order-lambda1=lambda2=0}
\left|u(t) - u^*(t)\right| \leq \lim_{\lambda_1, \lambda_2 \to 0}\varepsilon\phi(t;\lambda_1, \lambda_2) = \frac{\varepsilon t^2}{2} = O(\varepsilon t^2).
\end{equation}
A detailed derivation of Eq. \ref{eq:2nd-order-lambda2=0} and Eq. \ref{eq:2nd-order-lambda1=lambda2=0} can be found in Appendex \ref{appendix:2nd-order-limit-derivation}.



%
%
%

\subsection{System of First-Order Linear ODEs with Constant Coefficients} \label{section:ode-system}
\textbf{Proposition} \quad 
Let the $p$-norm of the residual $\left\|R\mathbf{u}(t)\right\|$ of the linear system $\mathbf{u}' + A \mathbf{u} = \mathbf{f}$ ($\mathbf{u}, \mathbf{f} \in \mathbb{C}^n$ and $A \in \mathbb{C}^{n\times n}$) be bounded by some $\varepsilon \geq 0$ on $I$, namely, 
\begin{equation}\label{eq:system}
    \left\| \mathbf{u}' + A \mathbf{u} - \mathbf{f} \right\| \leq \varepsilon \quad \forall t \in I,
\end{equation}
and the network solution satisfy the initial condition $\mathbf{u}(t_0) = \mathbf{u}^*_0$. Denote the Jordan canonical form of $A$ as
\begin{equation}
\resizebox{\hsize}{!}{$
    J = M^{-1}AM = \begin{pmatrix} J_1 & & & \\ & J_2 & & \\ & & \ddots & \\ & & & J_m \end{pmatrix} \text{ where } 
    J_k = \begin{pmatrix} \lambda_k + i\omega_k & 1 & & \\ & \ddots & \ddots & \\ & & \lambda_k + i\omega_k & 1 \\ & & & \lambda_k + i\omega_k \end{pmatrix} \quad k = 1, \dots, m
$}
\end{equation}
where $M$ is composed of generalized eigenvectors and $J_k$ ($1\leq k\leq m \leq n$) is a $n_k \times n_k$ Jordan block ($n_1 + \dots + n_m = n$). Then, the absolute error is bounded by $\| \mathbf{u} - \mathbf{u}^* \| \leq O(\varepsilon)$ if $\lambda_k > 0$ for all $k$.

\textbf{Proof} \quad 
With the substitution $\mathbf{v} := M^{-1} \mathbf{u}$, $\mathbf{g} := M^{-1} \mathbf{f}$, Eq. \ref{eq:system} can be transformed into 

\begin{equation}\label{eq:system-decoupled}
    \left\| \mathbf{v}' + J \mathbf{v} - \mathbf{g} \right\| = \left\| M^{-1} \mathbf{u}' + M^{-1}A \mathbf{u} - M^{-1}\mathbf{f} \right\| \leq \left\|M^{-1}\right\| \left\|\mathbf{u}' +A \mathbf{u} - \mathbf{f} \right\| \leq \left\|M^{-1}\right\| \varepsilon
\end{equation}

where $\left\|M^{-1}\right\|$ is the induced $p$-norm of $M^{-1}$. Each entry in $\left(\mathbf{v}' + J\mathbf{v} - \mathbf{g}\right)$ must be no greater than $\left\|M^{-1}\right\|\varepsilon$ in order for Eq. \ref{eq:system-decoupled} to hold. To bound the error for each Jordan chain, we first define two auxiliary sequence of functions $\{h_k\}$ and $\{H_k\}$, which will be useful in following derivations.
\begin{equation}\label{eq:h-H-definition}
    h_k(t;\lambda) := \frac{1}{\lambda^k}\left(1 - \sum_{j=0}^{k-1} \frac{\lambda^j (t-t_0)^j}{j!} e^{\lambda (t_0 - t)} \right)
    \quad\text{and}\quad
    H_k(t;\lambda) := \sum_{j=1}^{k} h_k(t;\lambda).
\end{equation}

Notice the property that, if $\lambda > 0$
$$ 0 \leq h_k(t;\lambda) < \frac{1}{\lambda^k} \quad 0 \leq H_k(t;\lambda) < \sum_{j=1}^{k} \frac{1}{\lambda^j} \quad \forall t \in I.$$ 

Now, consider the first Jordan chain, 
\begin{align}
    \left|v'_{1} + (\lambda_1 + i\omega_1) v_{1} + v_{2} - g_{1}\right| &\leq \left\|M^{-1}\right\|\varepsilon \\
    \vdots \nonumber \\
    \left|v'_{n_1 - 1} + (\lambda_1 + i\omega_1) v_{n_1 - 1} + v_{n_1} - g_{n_1 - 1}\right| &\leq \left\|M^{-1}\right\|\varepsilon \label{eq:system-jordan-second2last}\\
    \left|v'_{n_1} + (\lambda_1 + i\omega_1) v_{n_1} - g_{n_1}\right| &\leq \left\|M^{-1}\right\|\varepsilon \label{eq:system-jordan-last}
\end{align}

If $\lambda_1 > 0$, Eq. \ref{eq:system-jordan-last} implies (by section \ref{section:1st-order}) the absolute error bound on $v_{n_1}$
\begin{equation}\label{eq:system-jordon-last-bound}
    \left| v_{n_1} - v^*_{n_1} \right| \leq \left\|M^{-1}\right\| \varepsilon \frac{1 - e^{\lambda_1 (t_0 - t)}}{\lambda_1} = H_1(t; \lambda_1)\left\|M^{-1}\right\| \varepsilon
\end{equation}

Plugging Eq. \ref{eq:system-jordan-second2last} and Eq. \ref{eq:system-jordon-last-bound} into the following triangle inequality yields
\begin{align}
    \left|v'_{n_1 - 1} + (\lambda_1 + i\omega_1) v_{n_1 - 1} + v^*_{n_1} - g_{n_1 - 1}\right| &\leq \left|v'_{n_1 - 1} + (\lambda_1 + i\omega_1) v_{n_1 - 1} + v_{n_1} - g_{n_1 - 1}\right| + \left| v^*_{n_1} - v_{n_1}\right| \nonumber \\
    &\leq \left\|M^{-1}\right\|\varepsilon + H_1(t; \lambda_1)\left\|M^{-1}\right\| \varepsilon \label{eq:system-jordan-second2last-loose}
\end{align}
Apply the integrating factor technique again, there is
\begin{equation}
    \left|v_{n_1 - 1} - v^*_{n_1 - 1} \right| \leq H_2(t;\lambda) \left\|M^{-1}\right\| \varepsilon
\end{equation}

Repeating the above procedure, there is
\begin{equation}\label{eq:system-lambda>0}
\resizebox{.9\hsize}{!}{$
    \left|v_{1} - v^*_{1} \right| \leq H_{n_1}(t;\lambda_1) \left\|M^{-1}\right\| \varepsilon ,\quad
    \left|v_{2} - v^*_{2} \right| \leq H_{n_1-1}(t;\lambda_1) \left\|M^{-1}\right\| \varepsilon ,\quad
    \dots \quad,
    \left|v_{n_1} - v^*_{n_1} \right| \leq H_{1}(t;\lambda_1) \left\|M^{-1}\right\| \varepsilon
$}
\end{equation}

If $\lambda_1 = 0$, it can be proven (see Appendix \ref{appendix:system-lambda=0-proof}) that
\begin{equation}\label{eq:system-lambda=0}
\resizebox{.9\hsize}{!}{$
    \left|v_{1} - v^*_{1} \right| \leq \left\|M^{-1}\right\| \varepsilon \sum\limits_{j = 1}^{n_1} \frac{(t - t_0)^{j}}{j !}, \quad
    \left|v_{2} - v^*_{2} \right| \leq \left\|M^{-1}\right\| \varepsilon \sum\limits_{j = 1}^{n_1 - 1} \frac{(t - t_0)^{j}}{j !},\quad
    \dots \quad,
    \left|v_{n_1} - v^*_{n_1} \right| \leq \left\|M^{-1}\right\| \varepsilon (t - t_0)
$}
\end{equation}
 

Similarly, if $\lambda_k > 0$ for the $k$-th Jordan chain, then
\begin{align*}
    \left|v_{n_1 + \dots + n_{k-1} + 1} - v^*_{n_1 + \dots + n_{k-1} + 1} \right| &\leq H_{n_k}(t; \lambda) \left\|M^{-1}\right\| \varepsilon\\
    \left|v_{n_1 + \dots + n_{k-1} + 2} - v^*_{n_1 + \dots + n_{k-1} + 2} \right| &\leq H_{n_k - 1}(t; \lambda) \left\|M^{-1}\right\| \varepsilon\\
    \vdots \\
    \left|v_{n_1 + \dots + n_{k-1} + n_k} - v^*_{n_1 + \dots + n_{k-1} + n_k} \right| &\leq H_{1}(t; \lambda) \left\|M^{-1}\right\| \varepsilon
\end{align*}

It can be shown that, if $\lambda_k > 0$ for all $k$, then
\begin{equation} \label{eq:system-v-bound}
    \left\| \mathbf{v} - \mathbf{v}^* \right\| \leq \sqrt[p]{n} \left(\max_{k} \sum_{j=1}^{n_k} \frac{1}{\lambda_k^j} \right) \left\|M^{-1}\right\| \varepsilon.
\end{equation}
Substituting $\mathbf{u} = M\mathbf{v}$ into Eq. \ref{eq:system-v-bound}, we have the absolute error bound on $\mathbf{u}$,
\begin{equation} \label{eq:system-u-bound}
    \left\| \mathbf{u} - \mathbf{u}^* \right\| = \left\| M\mathbf{v} - M\mathbf{v}^* \right\| \leq \|M\| \left\|\mathbf{v} - \mathbf{v}^* \right\| 
    \leq \sqrt[p]{n} \left(\max_{k} \sum_{j=1}^{n_k} \frac{1}{\lambda_k^j} \right)\mathrm{cond}(M)\varepsilon = O(\varepsilon)
\end{equation}
where $\mathrm{cond}(M)= \left\|M\right\|\left\|M^{-1}\right\|$ is the condition number of $M$. Note that the matrix of generalized eigenvectors, $M$, can be replaced with $MD$ where $D\in \mathbb{C}^{n\times n}$ is a diagonal matrix. The infimum of condition number under right multiplication
$$\mathrm{cond^{R}}(M) := \inf_{D \text{ diagonal}} \mathrm{cond}(MD) = \inf_{D\text{ diagonal}} \|MD\|\left\|D^{-1}M^{-1}\right\|$$
has been studied for induced $1$-norm, $2$-norm, and $\infty$-norm in \cite{bauer1961absolute}, \cite{bauer1963optimally}, and \cite{braatz1994minimizing}.

\subsection{First-Order Linear ODE with Nonconstant Coefficients} \label{section:nonconstant}
\textbf{Proposition} \quad
Let the residual $\left|Ru(t)\right|$ of $u' + \left(p(t) + iq(t)\right) u = f(t)$ ($p, q: I \to \mathbb{R}, f: I\to \mathbb{C}$) be bounded by some $\varepsilon \geq 0$ on $I$, namely, 
\begin{equation}\label{eq:nonconstant-1st-order}
    \left| u' + \left(p(t) + i q(t)\right) u - f(t) \right| \leq \varepsilon \quad \forall t \in (t_0, \infty),
\end{equation}
and the network satisfy the initial condition $u(t_0) = u^*_0$, then the absolute error is bounded by 
\begin{equation}\label{eq:nonconstant-bound}
    \left| u - u^* \right| \leq O(\varepsilon t)
\end{equation}
if $p(t) \geq 0$ for sufficiently large $t$ on $I$.

\textbf{Proof} \quad 
Denote the antiderivatives of $p(t)$ and $q(t)$ as $$P(t) = \int_{t_0}^{t} p(\tau)\mathrm{d}\tau \quad Q(t) = \int_{t_0}^{t} q(\tau)\mathrm{d}\tau.$$
Applying the integrating factor technique again, there is

\begin{multline*}
    \left| \int_{t_0}^{t}  e^{P(\tau) + iQ(\tau)} \Big( u'(\tau) + \big(p(\tau) + i q(\tau)\big) u(\tau)  - f(\tau)\Big)\mathrm{d}\tau \right|  \\
    \leq \int_{t_0}^{t} \left| e^{P(\tau) + iQ(\tau)} \right| \left| u'(\tau) + \big(p(\tau) + i q(\tau)\big) u(\tau)  - f(\tau)\right|  \mathrm{d}\tau
\end{multline*}
\begin{align}
    \left| e^{P(t) + iQ(t)} u(t) - u(t_0) - \int_{t_0}^{t} e^{P(\tau) + iQ(\tau)} f(\tau) \mathrm{d}\tau \right| &\leq \varepsilon \int_{t_0}^{t} e^{P(\tau)} \mathrm{d}\tau \nonumber \\
    \left| u(t) - e^{- P(t) - iQ(t)}u^*_0 - e^{- P(t) - iQ(t)} \int_{t_0}^{t} e^{P(\tau) + iQ(\tau)} f(\tau)\mathrm{d}\tau \right| &\leq \varepsilon e^{-P(t)} \int_{t_0}^{t} e^{P(\tau)} \mathrm{d}\tau \nonumber \\
    \left| u(t) - u^*(t) \right| &\leq \varepsilon e^{-P(t)} \int_{t_0}^{t} e^{P(\tau)} \mathrm{d}\tau \label{eq:nonconstant-unclear}.
\end{align}

Rewriting the R.H.S. of of Eq. \ref{eq:nonconstant-unclear}, there is
\begin{equation}
    \left| u(t) - u^*(t) \right| \leq \varepsilon t \left(1 + \cfrac{\phi(t)}{te^{P(t)}}\right) ,
\end{equation}
where
\begin{equation}
    \phi(t) = \int_{t_0}^{t} e^{P(\tau)} \mathrm{d}\tau -te^{P(t)} = \int_{t_0}^{t} \left(e^{P(\tau)} - e^{P(t)}\right) \mathrm{d}\tau.
\end{equation}
Let $p(t) \geq 0$ for $t > t'$. Subsequently, $P(t)$ is nondecreasing for $t > t'$. Therefore, 

\begin{equation}
    \phi(t) = \int_{t_0}^{t'} \left(e^{P(\tau)} - e^{P(t)}\right) \mathrm{d}\tau + \int_{t'}^{t} \left(e^{P(\tau)} - e^{P(t)}\right) \mathrm{d}\tau \leq  \int_{t_0}^{t'} \left(e^{P(\tau)} - e^{P(t)}\right) \mathrm{d}\tau = \phi(t') \quad t > t'.
\end{equation}

Consequently, 
\begin{equation}
    \frac{\phi(t)}{te^{P(t)}} \leq \max_{\tau \in [t_0, t']} \left[\frac{\phi(\tau)}{\tau e^{P(\tau)}}\right] =: M,
\end{equation}
and finally,
\begin{equation}
    \left| u(t) - u^*(t) \right| \leq \varepsilon t \left(1 + M\right) = O(\varepsilon t).
\end{equation}

\subsection{Dividing the Intervals for a Tightened Error Bound} \label{section:subintervals}
In Sections \ref{section:1st-order} to \ref{section:nonconstant}, we only consider the global maximum residual norm $\varepsilon$ on $I$. However, one can also partition $I$ into subintervals $I = I_1 \cup I_2 \cup \dots$ and consider the local maximum residual norm $\varepsilon_k$ on $I_k$. This leads to an even tighter error bound since $\varepsilon_k \leq \varepsilon$ for all $k$.

For instance, in the case for first-order linear ODE with constant coefficients, the bound in Eq. \ref{eq:1st-order-abs} becomes 

\begin{equation}
    \left| u - u^* \right| \leq e^{-\lambda t} \int_{t_0}^t e^{\lambda \tau} \left|Ru(\tau)\right| \mathrm{d}\tau 
\end{equation}

as $\max\limits_{k}\rho(I_k) \to 0$, where $\rho(I_k)$ is the diameter of interval $I_k$.


\section{Experimental Results} \label{section:experiments}

For any ODE (or system of ODEs) discussed in Section \ref{section:error-bound-proof-and-methodology}, we are able to bound the error of any network by simply evaluating its infinite norm (maximum residual). This is true for any neural network solution, regardless of how well it is trained or trained at all. In addition, the error bound can be tightened by dividing the domain into subintervals and evaluating the infinite norm on each one.

We run the following experiments with the NeuroDiffEq library \cite{chen2020neurodiffeq}, which provides a convenient and flexible framework for training neural networks to solve differential equations. Unless otherwise specified, we use an Adam optimizer with a learning rate of $1.0 \times 10 ^{-3}$ and $(\beta_1, \beta_2)= (0.9,\, 0.999)$ for training the networks. The neural networks are simple fully-connected neural networks, with two 32-unit hidden layers and $\tanh$ activation function. The loss function we use is the $L_2$-norm of the ODE residuals at sampled points in the domain. The solution we choose is the one from the epoch with the lowest validation loss. We apply the following reparametrization to enforce the initial conditions $\mathbf{u}(t_0) = \mathbf{u}_0$ and, where required, $\cfrac{\mathrm d}{\mathrm{d}t}\mathbf{u}(t_0) =\mathbf{u}'_0$:

\begin{align}
    \mathbf{u}(t) &= \mathbf{u}_0 + \left(1 - e^{-(t-t_0)}\right) \mathrm{ANN}(t), \\
    \text{or}\quad\mathbf{u}(t) &= \mathbf{u}_0 + \left(t - t_0\right)\mathbf{u}'_0 + \left(1 - e^{-\left(t-t_0\right)^2}\right) \mathrm{ANN}(t).
\end{align}

\subsection{First-Order Linear ODE with Constant Coefficients} \label{section:1st-order-experiment}

We consider the real-valued ODE with constant coefficients, under the initial condition $u(0) = 2$,
\begin{equation*}
    u' + 3u = f(t) \quad t\in I=[0, 3]
\end{equation*}
for  $f(t) = 3t^2+5t+4$, $f(t) = 6\cos 3t$, $f(t) = 4e^t$, and $f(t) = -9\ln(1+t) - (1-t)^{-2}$,
where the exact solutions are $u(t) = e^{-3t} + t^2 + t + 1$, $u(t) = e^{-3t} + \sin 3t + \cos 3t$, $u(t) = e^{-3t} + e^{t}$, and $u(t) = e^{-3t} - 3 \ln(t+1) + (1 + t)^{-1}$. By Eq. \ref{eq:1st-order-abs}, the error bound for a single interval is given by $\varepsilon(1 - e^{-3t})/3$ where $\varepsilon$ is the largest absolute residual over the interval. 

Here we adopt the method of manufactured solutions for ease of computing absolute error. Namely, we pick specific forcing terms $f(t)$ so that the exact solutions are known. We then compute the absolute errors of neural networks using the known exact solutions. 

\begin{figure}
    \centering
    \includegraphics[width=0.95\textwidth]{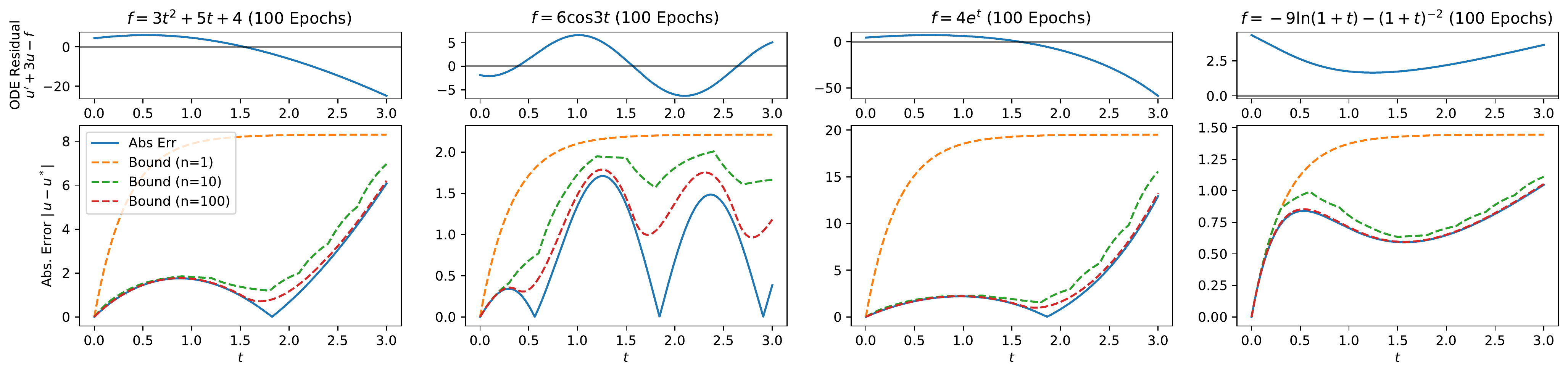}
    \includegraphics[width=0.95\textwidth]{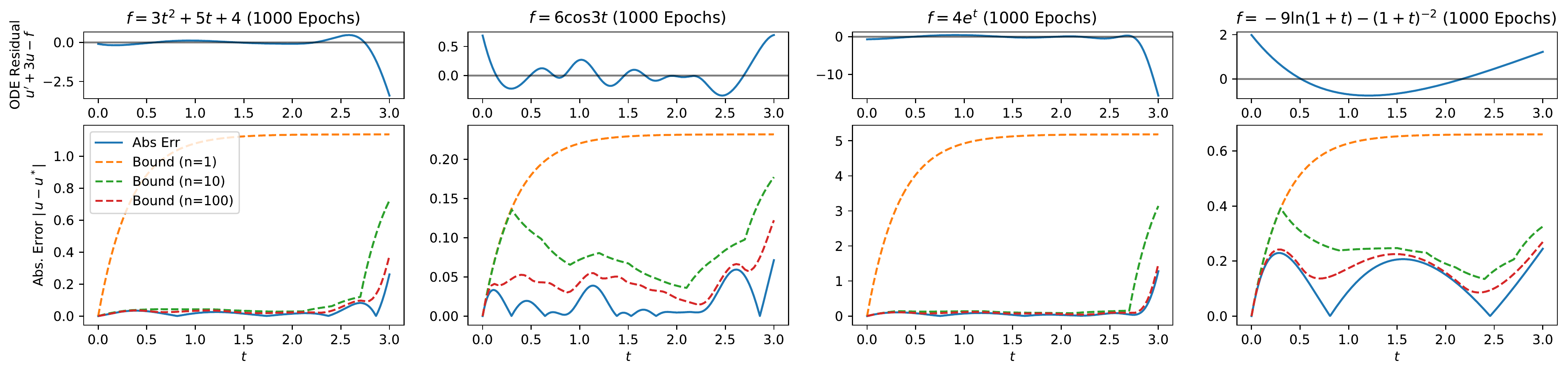}
    \caption{ODE Residuals and Absolute Error of ANN solution of Stable First Order System Under Forcing. The absolute error is calculated using the exact solution.}
    \label{fig:1st-order-const-interval}
\end{figure}

For each epoch, we uniformly sample 1024 points from the domain $I=[0, 3]$ for training. After $100$ and $1000$ epoch of training, we evaluate the residual of the ANN solution over $I$ and derive the corresponding error bounds using the propositions in Sections \ref{section:1st-order} and \ref{section:subintervals}. In Figure \ref{fig:1st-order-const-interval}, we evaluate the error bounds by linearly dividing the domain $I$ into $1$, $10$, $100$ sub-intervals and the corresponding error bounds are tighter as more sub-intervals are used. The absolute error computed using exact solution consistently fall within our derived bound, which verifies our approach for evaluating error bound.

\subsection{Higher-Order Linear ODE with Constant Coefficients}  \label{section:higher-order-experiment}
Here we consider two types of second-order differential equation, 
\begin{equation*}
    u'' + u = f \quad \text{and} \quad u'' + 4u' + 3u = f
\end{equation*}
where the solution space of the the associated homogeneous solution has basis $\{\sin t, \cos t\}$ and $\{e^{-t}, e^{-3t}\}$ respectively. By Eq. \ref{eq:2nd-order-lambda1=lambda2=0} and \ref{eq:2nd-order-bound-master} the error bounds for a single interval are $\varepsilon t^2/2$ and $\varepsilon\left(2 + e^{-3 t} - 3 e^{- t}\right)/6 \leq \varepsilon/3$ respectively, where $\varepsilon$ is the largest absolute residual over the interval. 

\begin{figure}
    \centering
    \includegraphics[width=0.95\textwidth]{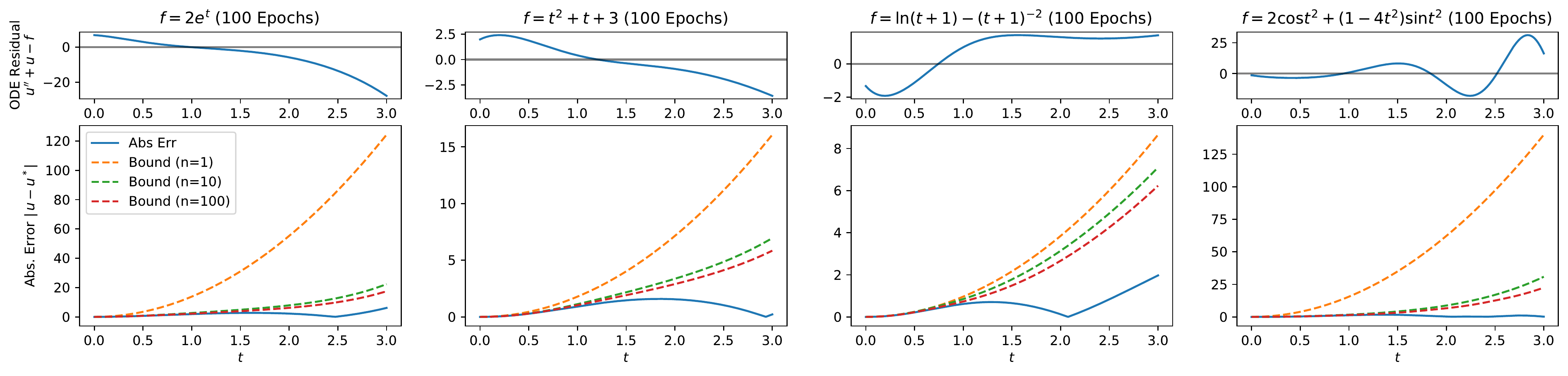}
    \includegraphics[width=0.95\textwidth]{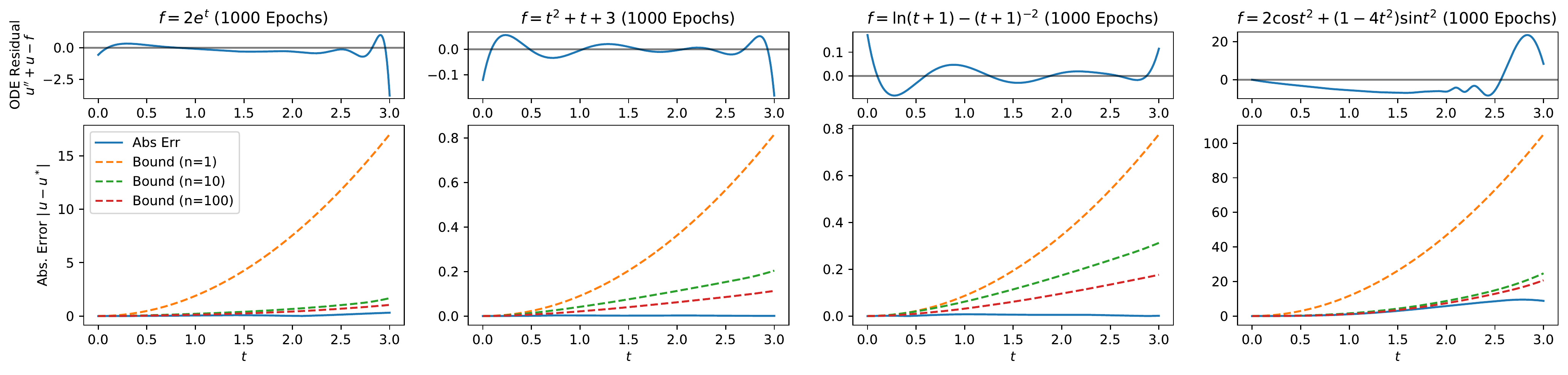}
    \caption{Residuals and Error Bounds of Harmonic Oscillator Under Various Forcing }
    \label{fig:2nd-order-oscillator}
\end{figure}

\begin{figure}
    \centering
    \includegraphics[width=0.95\textwidth]{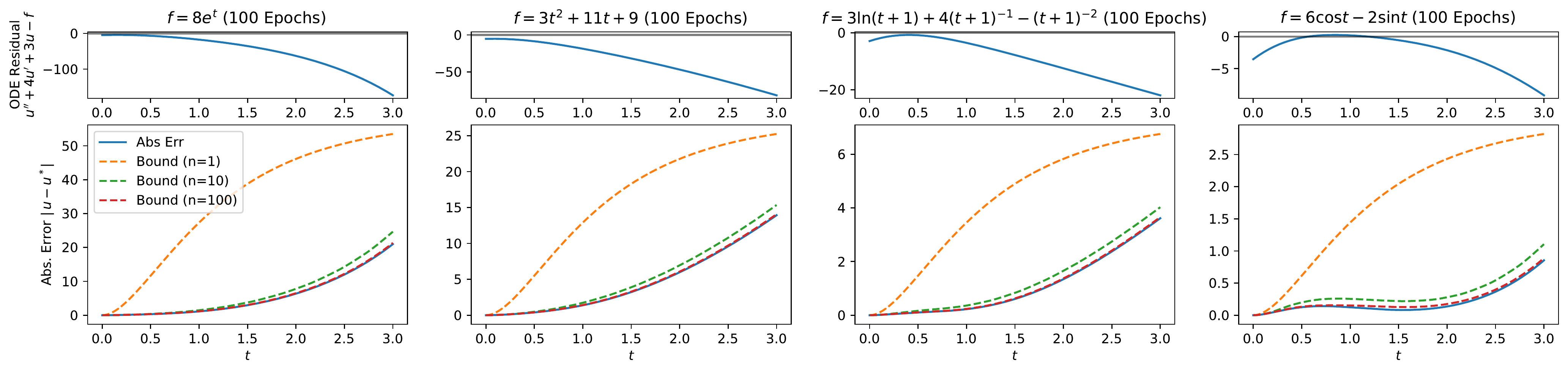}
    \includegraphics[width=0.95\textwidth]{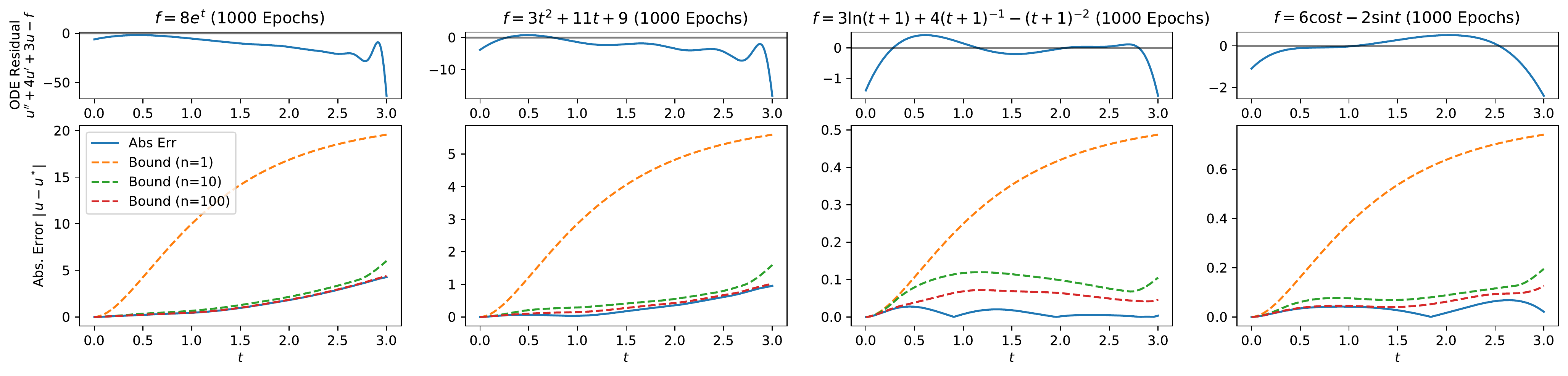}
    \caption{Residuals and Error Bounds of Stable 2nd Order Equation Under Various Forcing}
    \label{fig:2nd-order-exponential}
\end{figure}

We pick the forcing terms and initial conditions as described in Table \ref{tab:higher-order-constant}.
\begin{table}[]
    \centering
    \resizebox{\textwidth}{!}{
        \begin{tabular}{|c|c|c|c|l|}
            \hline
            Equation & Forcing $f(t)$& $u(0)$ & $u'(0)$ & Exact Solution $u(t)$ \\
            \hline
             $u''+u=f$ & $2e^t$ & $2.0$ & $2.0$ & $\sin t + \cos t + e^t$ \\
             $u''+u=f$ & $t^2+t+3$ & $2.0$ & $2.0$ & $\sin t + \cos t +t^2+t+1$ \\
             $u''+u=f$ & $\ln (t+1) - (t+1)^{-2}$ & $1.0$ & $2.0$ & $\sin t + \cos t + \ln(t+1)$ \\
             $u''+u=f$ & $2 \cos t^2 + (1-4t^2) \sin t^2$ & $1.0$ & $1.0$ & $\sin t + \cos t+ \sin t^2$ \\
             $u''+3u'+4u=f$ & $8e^t$ & $3.0$ & $-3.0$ & $e^{-t} + e^{-3t} + e^t$ \\
             $u''+3u'+4u=f$ & $3t^2+11t+9$ & $3.0$ & $-3.0$ & $e^{-t} + e^{-3t} + t^2+t+1$ \\
             $u''+3u'+4u=f$ & $3\ln(t+1) + 4(t+1)^{-1} - (t+1)^{-2}$ & $2.0$ & $-3.0$ & $e^{-t} + e^{-3t} + \ln(t+1)$ \\
             $u''+3u'+4u=f$ & $6\cos t - 2\sin t$ & $3.0$ & $-3.0$ & $e^{-t} + e^{-3t} + \sin t + \cos t$ \\
            \hline
        \end{tabular}
    }
    \caption{Experiment Setup for Section \ref{section:higher-order-experiment}}
    \label{tab:higher-order-constant}
\end{table}
We train the network on $I = [0, 3]$ for $100$ and $1000$ epochs. The ODE residual and error bound with $n=1, 10, 100$ subintervals are plotted in Figures \ref{fig:2nd-order-oscillator} and \ref{fig:2nd-order-exponential}.

\subsection{System of First-Order Linear ODEs with Constant Coefficients} \label{section:nonconstant-experiment}
\begin{figure}
    \centering
    \includegraphics[width=0.95\textwidth]{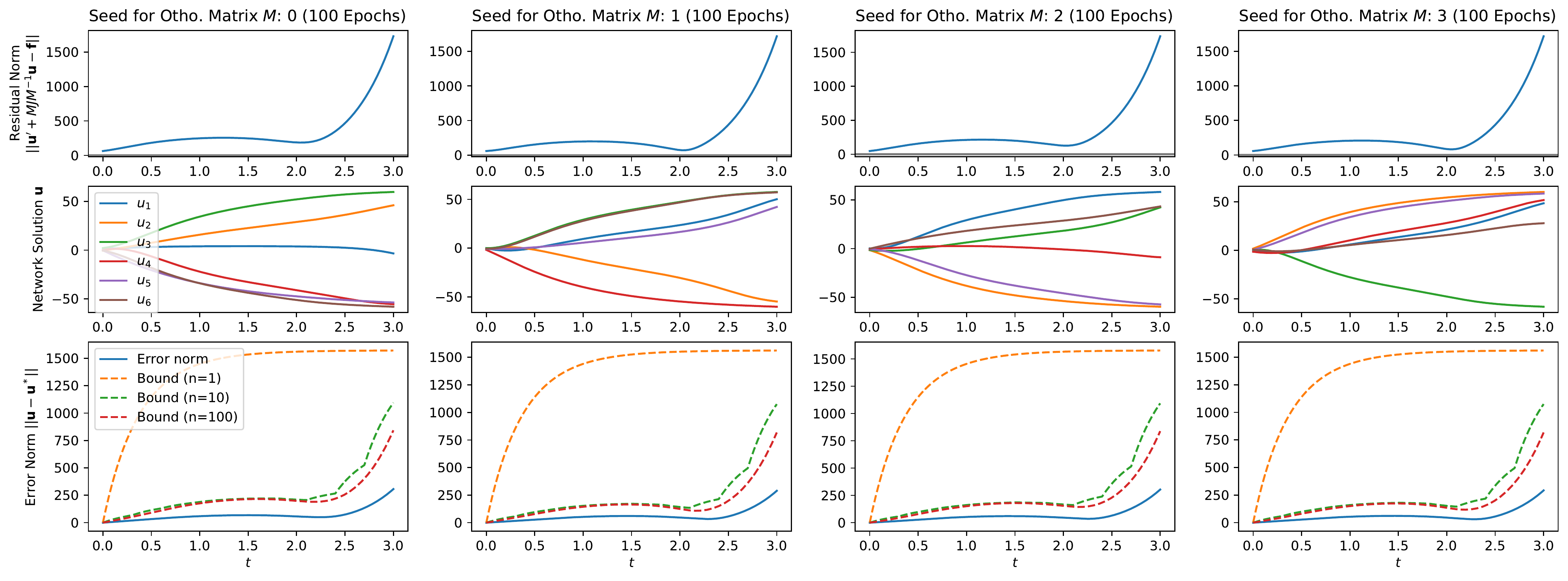}
    \includegraphics[width=0.95\textwidth]{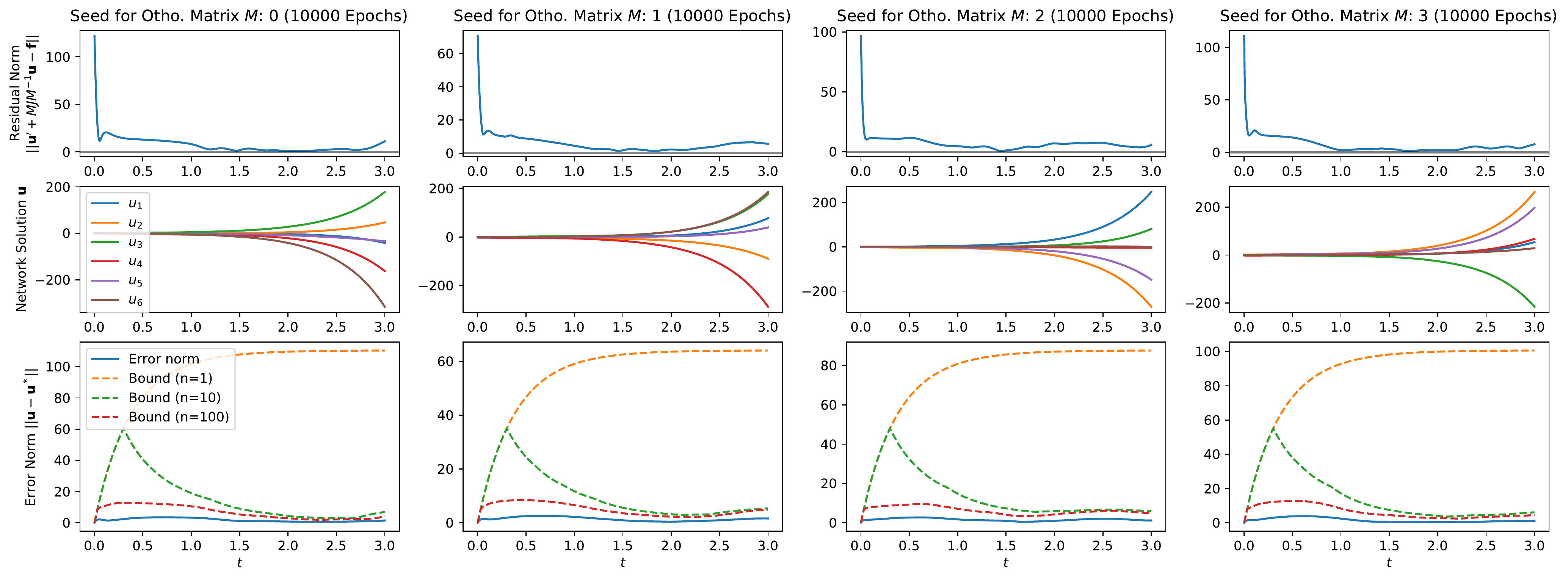}
    \caption{ODE Residuals and Absolute Error of ANN solution (System of Linear ODEs) }
    \label{fig:ode-system}
\end{figure}

We consider the following system of linear ODEs
\begin{equation*}
    \mathbf{u}' + MJM^{-1}\mathbf{u} = \mathbf{f}
\end{equation*} 
under the initial condition $\mathbf{u}(t_0) = M \begin{pmatrix}1&1&1&1&1&1\end{pmatrix}^T$ where $J\in \mathbb{R}^{6\times 6}$ is the Jordan canonical form $\begin{pmatrix}J_1\\ & J_2 \\ & & J_3\end{pmatrix}$ with Jordan blocks $J_1 = \begin{pmatrix}4&1\\&4&1\\&&4\end{pmatrix}$, $J_2 = \begin{pmatrix}3&1\\&3\end{pmatrix}$, and $J_3 = 2$, and the forcing is determined by
\begin{equation*}
    \resizebox{\hsize}{!}{$
    \mathbf{f}(t) = M\begin{pmatrix}
        \cos t + 4\sin t + e^t - 1 & 
        5 e^t - 4 + t^2 & 
        4t^2 + 2t & 
        3t^3+3t^2+e^{2t}-1 & 
        5e^{2t} - 3 & 
        \frac{1}{t+1} + 2 \ln(t+1)
        \end{pmatrix}^T
    $}
\end{equation*}
We randomly sample orthogonal $6\times 6$ matrices $M = M^{-T}$ to ensure $\mathrm{cond}(M) = 1$. 
By Eq. \ref{eq:system-lambda>0} and \ref{eq:system-u-bound}, the error bound is $\varepsilon\sqrt{H^2_3(t; 4)+H^2_2(t; 4) + H^2_1(t; 4) + H^2_2(t;3) + H^2_1(t;3) + H^2_1(t; 2)}$ $\leq \sqrt{6}\varepsilon/2$ for a single interval where $\varepsilon$ is the largest residual norm over the interval. 
The system is solved for $t \in I = [0, 3]$ for 1000 epochs with 1024 uniformly resampled points from the expanded domain $[-1, 4]$ at each epoch. We use networks with two 512-unit (instead of 32-unit) hidden layers due to the coupled nature of the system. However, it should be pointed out that, the error bound holds true regardless of the network size or how well the network is trained. Again, we divide $I$ into $n = 1, 10, 100$ subintervals for increasingly tighter bounds. Figure \ref{fig:ode-system} shows the system residual norm, network solution, as well as error bounds.

\subsection{First-Order Linear ODE with Nonconstant Coefficients}
In this section, we consider linear ODE with time-dependent coefficients, with $p(t)$, $f(t)$, initial condition and derived error bound for a single interval according to Eq. \ref{eq:nonconstant-unclear} tabulated in Table \ref{tab:first-order-nonconstant}.
\begin{equation}
    u' + p(t) u = f(t) \quad t \in I = [0, 3]
\end{equation}

\begin{table}[]
    \centering
    \resizebox{\textwidth}{!}{
        \begin{tabular}{|c|c|c|c|c|}
            \hline
            Forcing $f(t)$ & Coefficient $p(t)$  & IC $u(0)$ & Exact Solution $u(t)$ & Bound \\
            \hline
            $(2t+1)(t+1)^{-1}\cos t  - t\sin t$ & $(t+1)^{-1}$ & $1.0$ & $(t+1)^{-1} + t\cos t $ & $\varepsilon\frac{t^2/2+t}{t+1}  $\\
            $(t+1)^2\left(t^2 + 1\right)^{-1}e^{t}$ & $2t\left(t^2+1\right)^{-1}$ & $2.0$ & $(t^2+1)^{-1} + e^t$ & $\varepsilon\frac{t^3/3+t}{t^2+1}  $\\
            $2t+t^2\cos t\left(1 + \sin t\right)^{-1}$ & $\cos t\left(1+\sin t\right)^{-1}$ & $1.0$ & $\left(1+\sin t\right)^{-1} + t^2$ & $\varepsilon\frac{t-\cos t}{\sin t+1} $\\
            $\left(1+(t+2)\ln (t+1)\right)\left(t+1\right)^{-1} $ & $(t+2)(t+1)^{-1}$ & $1.0$ & $e^{-t}\left(t+1\right)^{-1} + \ln(t+1)$ & $\varepsilon\frac{e^t}{e^t+1} $\\
            \hline
        \end{tabular}
    }
    \caption{Experiment Setup $u'+p(t)u = f(t)$ for Section \ref{section:nonconstant-experiment}, where $\varepsilon$ in derived bounds is the largest absolute residual over a single interval}
    \label{tab:first-order-nonconstant}
\end{table}

\begin{figure}
    \centering
    \includegraphics[width=0.95\textwidth]{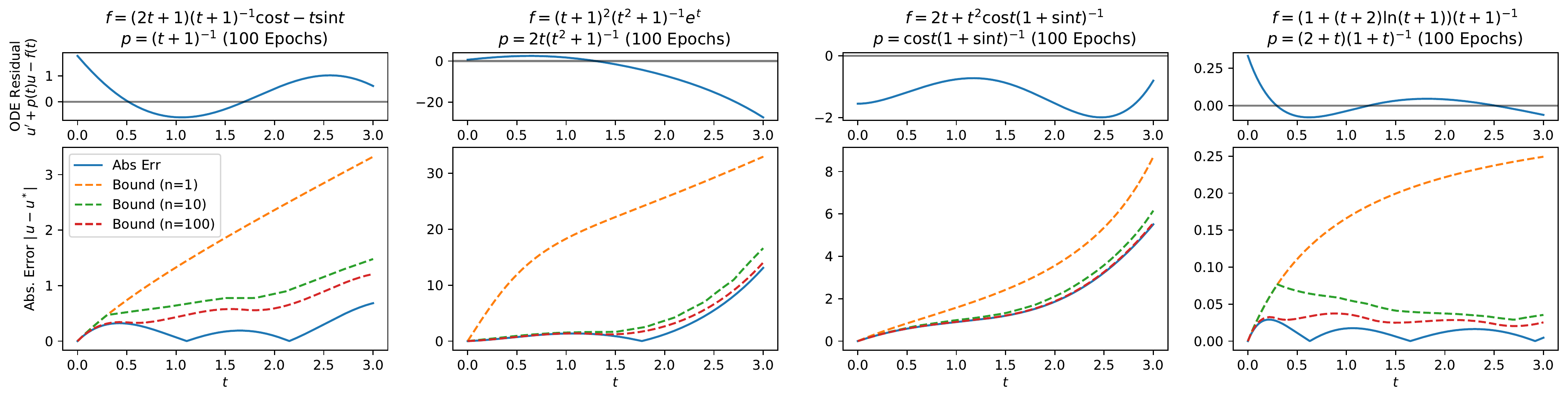}
    \includegraphics[width=0.95\textwidth]{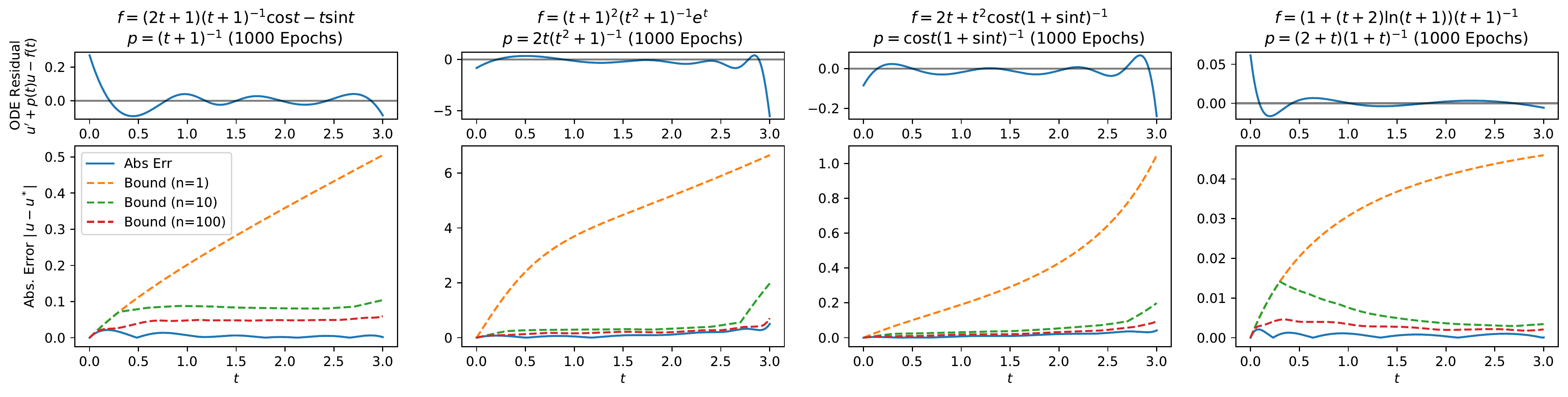}
    \caption{ODE Residuals and Absolute Error of ANN solution (Nonconstant Coefficients)}
    \label{fig:nonconstant}
\end{figure}

We train the network for 100 and 1000 epoch with 1024 points uniformly resampled from $I = [0, 3]$. According to Section \ref{section:nonconstant}, the absolute error bound is $O(\varepsilon t)$. By evenly dividing $I$ into $n = 1, 10, 100$ subintervals, we obtain the error bounds in Figure \ref{fig:nonconstant}.

\section{Conclusion} \label{section:conclusion}

In this work, we have ascertained the link between ODE residuals and error bound. We have proven, that for stable ODE systems discussed above, the bound only depends on characteristics of the ODE (or system of ODEs) and the maximum residual norm. There are efficient ways to evaluate the bound over the interval, as we did in Section \ref{section:experiments}.

In our experiments, we have shown that while these bounds are sometimes too loose using only the global maximum residual norm, they are usually asymptotically bounded by some constant. One can further tighten the bound by dividing the domain into smaller subintervals. In our experiments, the subintervals are linearly divided, but one can also use an adaptive quadrature for this task \cite{mckeeman1962algorithm}.

\section{Future Work} \label{section:future-work}

In this work, we tie linear ODEs residuals to the absolute error and showed the error can be bounded by a function of residuals. A subsequent research topic is what strategy can be used to ensure a low residual, which in turn guarantees a low absolute error.

As another future extension, our proposed method may be generalizable to system of linear ODEs with time-dependent coefficients. It is also interesting to study if this method is applicable to local linear approximation of nonlinear ODEs. Furthermore, spatial or spatiotemporal PDEs with Dirichlet boundary conditions is also worth exploring. 

\printbibliography

\appendix

\section{Derivation of Eq. \ref{eq:2nd-order-lambda2=0} and Eq. \ref{eq:2nd-order-lambda1=lambda2=0}} \label{appendix:2nd-order-limit-derivation}
Consider the case when $\lambda_1 > 0$, $\lambda_2 \to 0$, we have the following limit
\begin{align}
    \left|u-u^*\right| &\leq \lim_{\lambda_2 \to 0} \frac{\varepsilon}{\lambda_1 \lambda_2}\left(1 - \frac{\lambda_1 e^{-\lambda_2} - \lambda_2 e^{-\lambda_1}}{\lambda_1 - \lambda_2}\right) \nonumber \\
    &=\lim_{\lambda_2 \to 0} \frac{\varepsilon}{\lambda_1 \lambda_2} \frac{\lambda_1 - \lambda_2 - \left(\lambda_1 e^{-\lambda_2} - \lambda_2 e^{-\lambda_1}\right)}{\lambda_1 - \lambda_2} \nonumber \\
    &= \lim_{\lambda_2 \to 0} \frac{\varepsilon}{\lambda_1 \lambda_2} \frac{\lambda_1 \left(1 -e^{-\lambda_2 t}\right) - \lambda_2 \left(1 - e^{-\lambda_1 t}\right)}{\lambda_1 - \lambda_2} \nonumber \\
    &= \lim_{\lambda_2 \to 0} \frac{\varepsilon}{\lambda_1 \left(\lambda_1 - \lambda_2\right)} \frac{\lambda_1 \left(1 -e^{-\lambda_2 t}\right) - \lambda_2 \left(1 - e^{-\lambda_1 t}\right)}{\lambda_2} \nonumber \\
    &= \lim_{\lambda_2 \to 0} \frac{\varepsilon}{\lambda_1 \left(\lambda_1 - \lambda_2\right)} \left(\frac{\lambda_1 \left(1 -e^{-\lambda_2 t}\right)}{\lambda_2} - \left(1 - e^{-\lambda_1 t}\right) \right) \nonumber \\
    &= \frac{\varepsilon}{\lambda_1^2} \lim_{\lambda_2 \to 0} \left(\frac{\lambda_1 \left(1 -e^{-\lambda_2 t}\right)}{\lambda_2} - \left(1 - e^{-\lambda_1 t}\right) \right) \nonumber \\
    &= \frac{\varepsilon}{\lambda_1^2} \left(\lambda_1 t - 1 + e^{-\lambda_1 t} \right) \label{eq:2nd-order-lambda1-undecided} \\
    &\leq \frac{\varepsilon}{\lambda_1^2} \left(\lambda_1 t\right) = \frac{\varepsilon t}{\lambda_1} \nonumber
\end{align}

If we take the limit $\lambda_1 \to 0$ on top of $\lambda_2 \to 0$, step \ref{eq:2nd-order-lambda1-undecided} can be simplified using Taylor expansion,
\begin{align*}
    \left|u-u^*\right| &\leq \lim_{\lambda_1 \to 0} \frac{\varepsilon}{\lambda_1^2} \left(\lambda_1 t - 1 + e^{-\lambda_1 t} \right) \\
    &= \lim_{\lambda_1 \to 0} \frac{\varepsilon}{\lambda_1^2} \left(\lambda_1 t - 1 +  1 - \lambda_1 t + \cfrac{1}{2}\lambda_1^2 t^2 + t^3 O\left(\lambda_1^3\right) \right) \\
    &= \cfrac{\varepsilon t^2}{2}
\end{align*}

\section{General Case Proof of Section \ref{section:higher-order}} \label{appendix:higher-order-proof}

Define the following sequence of auxiliary functions $\{\phi_n\}_{n=1}^{\infty}$ on $I$,
\begin{equation*}
    \phi_n(t; \lambda_{1:n}) = \frac{1}{\prod\limits_{j=1}^{n} \lambda_j} - \sum_{k=1}^{n} \frac{e^{-\lambda_k (t-t_0)}}{\lambda_k\prod\limits_{j=1, j\neq k}^{n} \left(\lambda_j - \lambda_k\right)},
\end{equation*} 
where $\lambda_{1:n}$ is a tuple $\left(\lambda_1, \lambda_2, \dots, \lambda_n\right)$. 
Note that with $\phi_0(t) = 1$, it can be demonstrated that $\{\phi_n\}_{n=1}^{\infty}$ satisfies the recurrence relation
\begin{equation} 
    \phi_{n+1}(t; \lambda_{1:n+1}) = e^{-\lambda_{n+1} t} \int_{t_0}^{t} e^{\lambda_{n+1}\tau} \phi_n(\tau; \lambda_{1:n+1}) \mathrm{d} \tau \quad \text{for } n\geq 0.
\end{equation}

It can also be proven that $\phi_n(t;\lambda_{1:n})$ is monotonically increasing on $I$ if $\lambda_1,\dots,\lambda_n \geq 0$ because
\begin{equation}
    \frac{\mathrm d}{\mathrm d t}\phi_n(t, \lambda_{1:n}) = \sum_{k=1}^{n} \frac{e^{-\lambda_k (t-t_0)}}{\prod\limits_{j=1, j\neq k}^{n} \left(\lambda_j - \lambda_k\right)} \geq 0.
\end{equation}
Also, if $\lambda_1,\dots,\lambda_n > 0$, there is $\lim\limits_{t\to\infty} \phi_n(t, \lambda_{1:n}) = \prod\limits_{j=1}^{n}\lambda_j^{-1}$.

Let $u_0(t) := u(t)$, $u_1(t) := u'_0(t) + (\lambda_n + i\omega_n) u_0(t)$,  $u_2(t) := u'_1(t) + (\lambda_{n-1} + i\omega_{n-1}) u_1(t)$, \dots, $u_{n-1}(t) = u'_{n-2}(t) + (\lambda_{2} + i\omega_{2})u_{n-2}(t)$,
Eq. \ref{eq:higher-order-ode} can be written as 
\begin{equation}\label{eq:higher-order-reduced}
    \left|u_{n-1}' + (\lambda_1 + i\omega_1) u_{n-1} - f\right| \leq \varepsilon,
\end{equation}
which is a first-order inequality in terms of $u_{n-1}$ as discussed in Section \ref{section:1st-order}. By Eq. \ref{eq:1st-order-generic}, 
\begin{equation} \label{eq:higher-order-step1}
    \left|u_{n-1} - u^*_{n-1}\right| \leq \varepsilon e^{-\lambda_{1} t}\int_{t_0}^{t}e^{\lambda_{1}\tau}\mathrm{d}\tau = \varepsilon \phi_1(t; \lambda_1).
\end{equation}

Substitute $u_{n-1}(t) = u'_{n-2}(t) + (\lambda_{2} + i\omega_{2})u_{n-2}(t)$ back into Eq. \ref{eq:higher-order-step1}, we have 
\begin{equation*}
    \left|u_{n-2} + (\lambda_2 + i\omega_2) - u^*_{n-1}\right| \leq \varepsilon \phi_1(t; \lambda_1),
\end{equation*}
which is a first order inequality in terms of $u_{n-2}$. Applying the integrating factor trick again, we have 
\begin{equation}
    \left|u_{n-2} - u^*_{n-2}\right| \leq \varepsilon e^{-\lambda_2 t} \int_{t_0}^{t} e^{\lambda_2 t}\phi_1(\tau, \lambda_1) \mathrm{d} \tau = \varepsilon \phi_2(t; \lambda_{1:2}).
\end{equation}
Repeating the above process yields 

\begin{equation}
    \left|u - u^*\right| = \left|u_{0} - u^*_{0}\right| \leq \varepsilon \phi_n(t; \lambda_{1:n})
\end{equation}

\section{Proof of Equation \ref{eq:system-lambda=0}} \label{appendix:system-lambda=0-proof}
Take the limit $\lambda \to 0$ in Eq. \ref{eq:h-H-definition}, and applying Taylor expansions where necessary, we have
\begin{align*}
    h_k(t;0) &= \lim_{\lambda \to 0} \frac{1}{\lambda^k}\left(1 - \sum_{j=0}^{k-1} \frac{\lambda^j (t-t_0)^j}{j!} e^{\lambda (t_0 - t)} \right) \\
    &= \lim_{\lambda \to 0} \cfrac{e^{\lambda(t_0-t)}}{\lambda^k}\left(e^{\lambda(t-t_0)} - \sum_{j=0}^{k-1} \frac{\lambda^j (t-t_0)^j}{j!} \right) \\
    &= \lim_{\lambda \to 0} \cfrac{e^{\lambda(t_0-t)}}{\lambda^k}  \sum_{j=k}^{\infty} \frac{\lambda^j (t-t_0)^j}{j!} \\
    &= \lim_{\lambda \to 0} \cfrac{1}{\lambda^k} \left(\sum_{l=0}^{\infty} \frac{\lambda^{l} (t_0-t)^{l}}{l!}\right) \left(\sum_{j=k}^{\infty} \frac{\lambda^j (t-t_0)^j}{j!}\right) \\
\end{align*}
Notice the lowest order term w.r.t. $\lambda$ in $\displaystyle{\left(\sum_{l=0}^{\infty} \frac{\lambda^{l} (t_0-t)^{l}}{l!}\right) \left(\sum_{j=k}^{\infty} \frac{\lambda^j (t-t_0)^j}{j!}\right)}$ is $\lambda_k$, which is attained only when $l=0$ and $j=k$. The coefficient for the $\lambda^k$ term is given by
$$
    \cfrac{(t_0-t)^0}{0!} \cdot \cfrac{(t-t_0)^k}{k!} = \cfrac{(t-t_0)^k}{k!} .
$$
Consequently, 
\begin{align*}
    h_k(t;0) &= \lim_{\lambda \to 0} \cfrac{1}{\lambda^k} \left( \cfrac{(t-t_0)^k}{k!} \lambda^k + O\left(\lambda^{k+1}\right) \right) = \cfrac{(t-t_0)^k}{k!} \\
    H_k(t;0) &= \sum_{j=1}^{k} h_k(t;0) = \sum_{j=1}^{k} \cfrac{(t-t_0)^j}{j!}
\end{align*}
Eq. \ref{eq:system-lambda=0} is attained by plugging the above equality into Eq. \ref{eq:system-lambda>0}.

\section{Examples of dividing domain into subintervals}

In Section \ref{section:subintervals}, we show that the error bound on $I = [0, t]$ can be further tightened by evaluating the maximum absolute residuals on a sequence of subintervals $I_i = [t_{i-1}, t_i]$. We apply this technique for the experiments in Section \ref{section:experiments}. 

\subsection{Second Order Linear Equation with Constant Coefficients}

Consider a second-order linear equation with constant coefficients (assuming $\lambda_1, \lambda_2 \geq 0$,
\begin{equation} \label{eq:2nd-order-without-residual}
    u''(t) + (\lambda_1 + i\omega_1 + \lambda_2 + i\omega_2) u'(t) + (\lambda_1 + i\omega_1) (\lambda_2 + i\omega_2) u(t) = f(t)
\end{equation}

An approximated solution yielded by a neural network does not exactly satisfy the Eq. \ref{eq:2nd-order-without-residual}, but instead incurs what we call a residual term $r(t)$
\begin{equation} \label{eq:2nd-order-with-residual}
    u''(t) + (\lambda_1 + i\omega_1 + \lambda_2 + i\omega_2) u'(t) + (\lambda_1 + i\omega_1) (\lambda_2 + i\omega_2) u(t) = f(t) + r(t)
\end{equation}

Solutions of Eq. \ref{eq:2nd-order-with-residual} and \ref{eq:2nd-order-without-residual} differ by 
\begin{align*}
    \Delta(t) &= e^{-{\lambda_1} t} \int_{s=0}^{s=t}e^{{\lambda_1} s}e^{-{\lambda_2} s} \left( \int_{\tau = 0}^{\tau = s}e^{{\lambda_2} \tau} r(\tau) \mathrm{d}\tau\right)\mathrm{d}s \\
    &= e^{-{\lambda_1} t} \int_{s=0}^{s=t}e^{({\lambda_1} - {\lambda_2})s} \left( \int_{\tau = 0}^{\tau = s}e^{{\lambda_2} \tau} r(\tau) \mathrm{d}\tau\right)\mathrm{d}s \\
    &= e^{-{\lambda_1} t} \int_{s=0}^{s=t}\int_{\tau = 0}^{\tau = s}e^{({\lambda_1} - {\lambda_2})s} e^{{\lambda_2} \tau} r(\tau) \mathrm{d}\tau\mathrm{d}s
\end{align*}
Therefore
\begin{align*}
    |\Delta(t)| &\leq e^{-{\lambda_1} t} \int_{s=0}^{s=t}\int_{\tau = 0}^{\tau = s}e^{({\lambda_1} - {\lambda_2})s} e^{{\lambda_2} \tau} |r(\tau)| \mathrm{d}\tau\mathrm{d}s \\
    &= e^{-{\lambda_1} t} \int_{\tau = 0}^{\tau = t}\int_{s=\tau}^{s=t}e^{({\lambda_1} - {\lambda_2})s} e^{{\lambda_2} \tau} |r(\tau)| \mathrm{d}s \mathrm{d}\tau \\
    &= e^{-{\lambda_1} t} \int_{\tau = 0}^{\tau = t} e^{{\lambda_2} \tau} |r(\tau)| \left( \int_{s=\tau}^{s=t}e^{({\lambda_1} - {\lambda_2})s} \mathrm{d}s \right) \mathrm{d}\tau \\
    &= e^{-{\lambda_1} t} \int_{\tau = 0}^{\tau = t} e^{{\lambda_2} \tau} |r(\tau)| \frac{e^{({\lambda_1} - {\lambda_2})t}-e^{({\lambda_1} - {\lambda_2})\tau}}{{\lambda_1} - {\lambda_2}} \mathrm{d}\tau \\
    &= \int_{\tau = 0}^{\tau = t} |r(\tau)| \frac{e^{{\lambda_2}(\tau-t)}-e^{{\lambda_1}(\tau-t)}}{{\lambda_1} - {\lambda_2}} \mathrm{d}\tau
\end{align*}

Notice that $\cfrac{e^{{\lambda_2}(\tau-t)}-e^{{\lambda_1}(\tau-t)}}{{\lambda_1} - {\lambda_2}} \geq 0$ for $\tau < t$. Let $M(a,b) = \max\limits_{a\leq \tau \leq b} \left|r(\tau)\right|$, we have 
\begin{align}
    |\Delta(t)| &\leq \sum_{i=1}^{n}M(t_{i-1},t_i) \int_{\tau=t_{i-1}}^{\tau=t_{i}} \frac{e^{{\lambda_2}(\tau-t)}-e^{{\lambda_1}(\tau-t)}}{{\lambda_1} - {\lambda_2}} \mathrm{d}\tau \label{eq:2nd-order-subintervals} \\
    &\leq M_{(0,t)} \int_{\tau=0}^{\tau=t} \frac{e^{{\lambda_2}(\tau-t)}-e^{{\lambda_1}(\tau-t)}}{{\lambda_1} - {\lambda_2}} \mathrm{d}\tau \label{eq:2nd-order-single-interval}.
\end{align}
where $0 = t_0 < t_1 < \dots < t_n = t$.

Eq. \ref{eq:2nd-order-subintervals} sheds light on how to evaluate the error bound by subdividing interval $[0, t]$ into $n$ subintervals. Namely, we first evaluate the maximum absolute residual $M(t_{i-1}, t_i)$ on $[t_{i-1}, t_i]$ as well as the integral $\displaystyle{\int_{\tau=0}^{\tau=t} \frac{e^{{\lambda_2}(\tau-t)}-e^{{\lambda_1}(\tau-t)}}{{\lambda_1} - {\lambda_2}} \mathrm{d}\tau}$, which always has a closed-form expression depending $\lambda_1$ and $\lambda_2$. The absolute error at any $t$ is then bounded by the sum of the products. In particular, Eq. \ref{eq:2nd-order-single-interval} is the special case where we do not divide $[0, t]$ into subintervals ($n = 1$), which is discussed in Section \ref{section:higher-order}.

In the special case where $\max\limits_{1\leq i\leq n} (t_i - t_{i-1}) \to 0$, there is \begin{equation}
    \int_{\tau = 0}^{\tau = t} |r(\tau)| \frac{e^{{\lambda_2}(\tau-t)}-e^{{\lambda_1}(\tau-t)}}{{\lambda_1} - {\lambda_2}} \mathrm{d}\tau = \sum_{i=1}^{n}M(t_{i-1},t_i) \int_{\tau=t_{i-1}}^{\tau=t_{i}} \frac{e^{{\lambda_2}(\tau-t)}-e^{{\lambda_1}(\tau-t)}}{{\lambda_1} - {\lambda_2}} \mathrm{d}\tau . \label{eq:2nd-order-infinite-intervals}
\end{equation}



\subsection{System of ODEs}
Consider a Jordan chain of length $3$ and eigenvalue $(\lambda + i\omega)$.
\begin{align*}
    u'_1(t) + (\lambda +i \omega) u_1(t) + u_2(t) &= f_1(t) \\
    u'_2(t) + (\lambda +i \omega) u_2(t) + u_3(t) &= f_2(t) \\
    u'_3(t) + (\lambda +i \omega) u_3(t) &= f_3(t)
\end{align*}

The approximated solution given by the neural network incurs residuals $r_1(t), r_2(t), r_3(t)$, namely, 
\begin{align}
    u'_1(t) + (\lambda +i \omega) u_1(t) + u_2(t) &= f_1(t) + r_1(t) \label{eq:system-u1} \\
    u'_2(t) + (\lambda +i \omega) u_2(t) + u_3(t) &= f_2(t) + r_2(t) \label{eq:system-u2} \\
    u'_3(t) + (\lambda +i \omega) u_3(t) &= f_3(t) + r_3(t) \label{eq:system-u3}
\end{align}

Eq. \ref{eq:system-u3} implies that 
\begin{align*}
    |u_3 - u^*_3| \leq e^{-\lambda t} \int_{\tau=0}^{\tau=t} e^{\lambda \tau} |r_3(\tau)| \mathrm{d}\tau
\end{align*}

By triangle inequality, Eq. \ref{eq:system-u2} becomes
\begin{align*}
    \left|u'_2 + \lambda u_2 + u^*_3\right| &\leq \left|u'_2 + \lambda u_2 + u_3\right| + |u_3 - u^*_3| \leq |r_2(t)| + e^{-\lambda t} \int_{\tau=0}^{\tau=t} e^{\lambda t} |r_3(\tau)| \mathrm{d}\tau \\
    \left|u_2 - u^*_2\right|&\leq e^{-\lambda t} \int_{\tau=0}^{\tau=t} e^{\lambda \tau} |r_2(\tau)| \mathrm{d}\tau + e^{-\lambda t} \int_{s=0}^{s=t} \left( \int_{\tau=0}^{\tau=s} e^{\lambda \tau} |r_3(\tau)| \mathrm{d}\tau\right) \mathrm{d}s \\
    &= e^{-\lambda t} \int_{\tau=0}^{\tau=t} e^{\lambda \tau} |r_2(\tau)| \mathrm{d}\tau + e^{-\lambda t} \int_{\tau=0}^{\tau=t} \left( \int_{s=\tau}^{s=t}  e^{\lambda \tau} |r_3(\tau)| \mathrm{d}s\right) \mathrm{d}\tau \\
    &= e^{-\lambda t} \int_{\tau=0}^{\tau=t} e^{\lambda \tau} |r_2(\tau)| \mathrm{d}\tau + e^{-\lambda t} \int_{\tau=0}^{\tau=t} (t - \tau)e^{\lambda \tau} |r_3(\tau)|  \mathrm{d}\tau 
\end{align*}

Apply the same procedure for Eq. \ref{eq:system-u1}, there is
\begin{align*}
    \left|u'_1 + \lambda u_1 + u^*_2\right| &\leq \left|u'_1 + \lambda u_1 + u_2\right| + |u_2 - u^*_2| \\
    &\leq |r_1(t)| + e^{-\lambda t} \int_{\tau=0}^{\tau=t} e^{\lambda \tau} |r_2(\tau)| \mathrm{d}\tau + e^{-\lambda t} \int_{\tau=0}^{\tau=t} (t - \tau)e^{\lambda \tau} |r_3(\tau)|  \mathrm{d}\tau \\
    |u_1 - u^*_1| &\leq e^{-\lambda t}\int_{\tau=0}^{\tau=t} e^{\lambda \tau} |r_1(\tau)| \mathrm{d}\tau + e^{-\lambda t} \int_{s=0}^{s=\tau} \left( \int_{\tau=0}^{\tau=s} e^{\lambda \tau} |r_2(\tau)| \mathrm{d}\tau \right)\mathrm{d}s  \\
    &\quad + e^{-\lambda t} \int_{s=0}^{s=\tau} \left(\int_{\tau=0}^{\tau=s} (s - \tau)e^{\lambda \tau} |r_3(\tau)|  \mathrm{d}\tau\right)\mathrm{d}s \\
    &= e^{-\lambda t}\int_{\tau=0}^{\tau=t} e^{\lambda \tau} |r_1(\tau)| \mathrm{d}\tau + e^{-\lambda t} \int_{\tau=0}^{\tau=t} \left(\int_{s=\tau}^{s=t}  e^{\lambda \tau} |r_2(\tau)| \mathrm{d}s \right)\mathrm{d}\tau  \\
    &\quad + e^{-\lambda t} \int_{\tau=0}^{\tau=t} \left(\int_{s=\tau}^{s=t} (s - \tau)e^{\lambda \tau} |r_3(\tau)|  \mathrm{d}s\right)\mathrm{d}\tau \\
    &= e^{-\lambda t}\int_{\tau=0}^{\tau=t} e^{\lambda \tau} |r_1(\tau)| \mathrm{d}\tau + e^{-\lambda t} \int_{\tau=0}^{\tau=t} (t-\tau) e^{\lambda \tau} |r_2(\tau)|\mathrm{d}\tau + e^{-\lambda t} \int_{\tau=0}^{\tau=t} \frac{(t - \tau)^2}{2}e^{\lambda \tau} |r_3(\tau)|\mathrm{d}\tau \\
    &= \int_{\tau=0}^{\tau=t} e^{\lambda (\tau-t)} |r_1(\tau)| \mathrm{d}\tau + \int_{\tau=0}^{\tau=t} (t-\tau) e^{\lambda (\tau-t)} |r_2(\tau)|\mathrm{d}\tau + \int_{\tau=0}^{\tau=t} \frac{(t - \tau)^2}{2}e^{\lambda (\tau-t)} |r_3(\tau)|\mathrm{d}\tau
\end{align*}

Note that, with $M_k(a, b) = \max\limits_{a\leq\tau\leq b} |r_k(\tau)|$ ($k=1, 2, 3$) and $0 = t_0 \leq t_1 \leq \dots \leq t_n = t$,
\begin{align}
    0 \leq \int_{0}^{t} \frac{(t-\tau)^k}{k!}e^{\lambda (\tau - t)} |r(\tau)| \mathrm{d}\tau &\leq \sum_{i=1}^{n} M_k(t_{i-1}, t_i) \int_{t_{i-1}}^{t_i} \frac{(t-\tau)^k}{k!}e^{\lambda (\tau - t)} \mathrm{d}\tau \label{eq:system-subintervals} \\
    &\leq M_k(0, t) \int_{0}^{t} \frac{(t-\tau)^k}{k!} e^{\lambda(\tau -t)} \mathrm{d}\tau  \label{eq:system-single-interval}
\end{align}

Again, Eq. \ref{eq:system-subintervals} shows one can evaluate the absolute error bound by dividing $n$ subintervals. For each interval, one evaluates the maximum residual as well as the integral (which has a closed-form expression). Eq. \ref{eq:system-single-interval} is the special case as discussed in Section \ref{section:ode-system}, where subintervals are not used ($n=1$).


\end{document}